\documentclass[10pt]{article}

\usepackage{palmyra_neurips}
\usepackage{graphicx}

\title{Palmyra x6 Technical Report\\[4pt]
{\large An Agentic, Tool-Use Model Post-Trained via \\ Anchored Supervised Fine-Tuning}}
\author{%
  \textbf{Peng Du}\thanks{Corresponding authors: \texttt{dan.bikel@writer.com} and \texttt{peng@writer.com}.} \quad 
  \textbf{Kiran Kamble} \quad 
  \textbf{Rakshith Vasudev} \quad 
  \textbf{Zhizhuo Yang} \\
  \textbf{Rohith Nadimpally} \quad 
  \textbf{Arjun Krishna} \quad 
  \textbf{Waseem AlShikh} \\
  \textbf{Daniel M. Bikel\footnotemark[1]}\\
  \vspace{0.2cm} \\ 
  Writer AI Research, Writer, Inc.\\
  \texttt{\{peng, kiran, rakshith, zhizhuo, rohith, arjun,}\\
  \texttt{waseem, dan.bikel\}@writer.com}
}
\date{August 12, 2026}

\begin{document}

\maketitle


\section{Introduction}
\label{sec:intro}

Historically, in order to produce a large language model that was high-performing in many dimensions and excelled in a relatively small subset of interest, one would often take on the entire training pipeline, from pretraining, to continued pretraining/mid-training, to the various stages of post-training.  Recently, however, it often suffices to post-train an existing high-performing LLM, provided that the training data is of high quality and the objective is carefully chosen to match the needs of the model's intended applications. Palmyra~x6 is a case in point. The primary capability we are interested in is \emph{agentic task completion}: multi-step planning, tool/function calling, and long-horizon task completion within tool-rich environments such as MCP (Model Context Protocol) suites, web search stacks, code-execution sandboxes, and retrieval-augmented pipelines. After extensive testing, we chose a flagship open Mixture-of-Experts (MoE) base with 744B total and $\sim$40B active parameters per token. To it we apply a compact, high-quality corpus of synthetic agentic trajectories and an anchored fine-tuning objective~\citep{asft2025} designed to add agentic skill without eroding the abilities the base model already has.

\begin{figure}[h!]
\centering
\includegraphics[width=\linewidth]{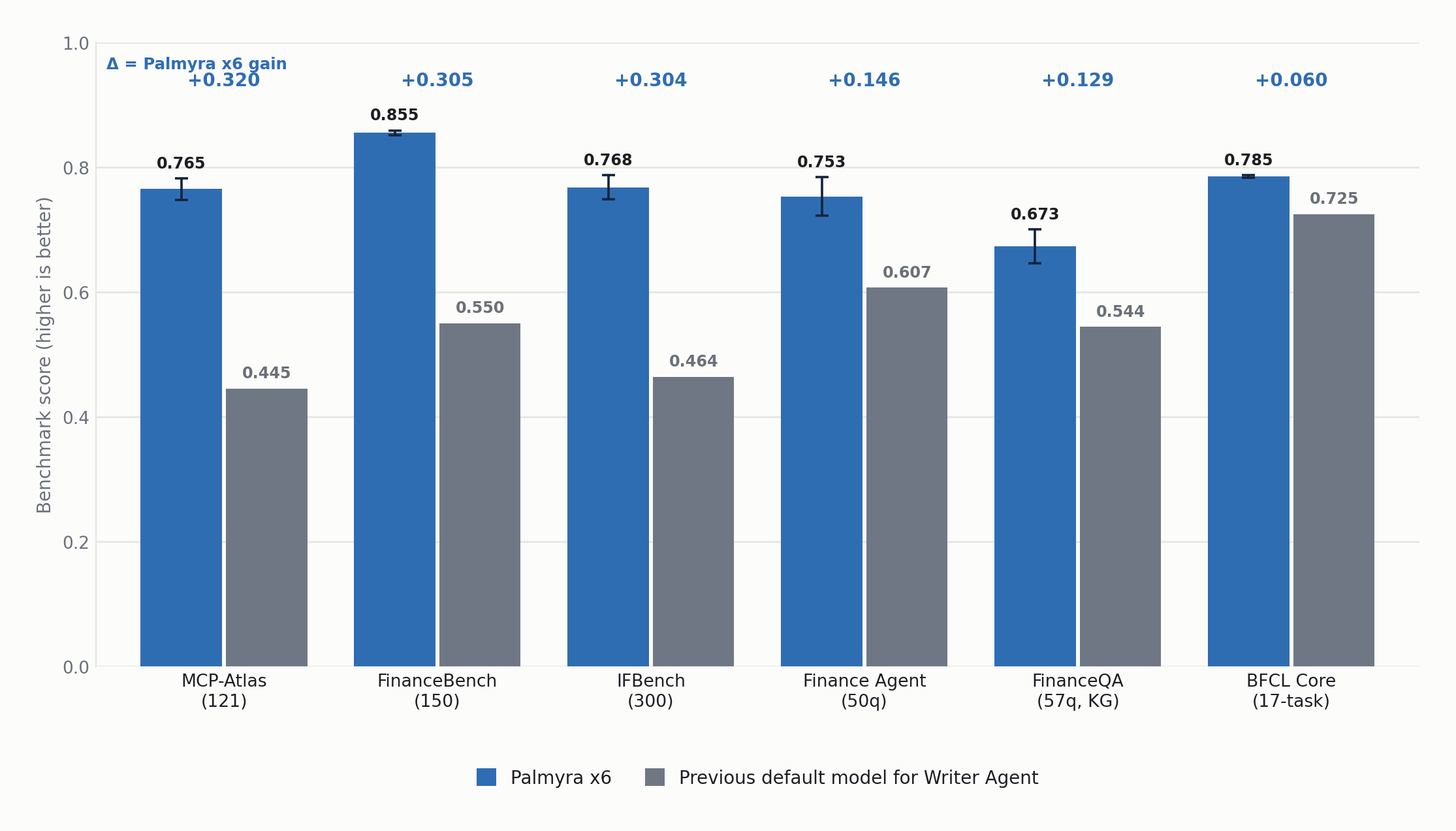}
\caption{Per-benchmark comparison of Palmyra x6 against the previous default model for Writer Agent on six evaluation benchmarks, ordered by margin. Bars show mean score (higher is better); error bars on Palmyra x6 denote one standard error. Palmyra x6 leads on every benchmark, with the largest gains on MCP-Atlas ($+0.320$), FinanceBench ($+0.305$), and IFBench ($+0.304$).}
\label{fig:eval-prior}
\end{figure}

In terms of task performance, Palmyra~x6 delivers a huge improvement over the previous default model for Writer Agent, as shown by Figure~\ref{fig:eval-prior}.  Comparisons against other frontier models available at the time of development are collected in Figure~\ref{fig:benchmarks}.  However, we note that we use public benchmark performance merely as a sanity check and a guide, but not as the ultimate rubric.  Rather, our focus was on developing internal datasets that were meant to mimic our enterprise customers' domains and use cases, and closely track model performance on those, as they are a much better reflection of how the model will do in production than public benchmarks.

\section{Overview}
\label{sec:overview}
Our model development proceeded in two phases:

\begin{enumerate}
  \item \textbf{Recipe development.} We ran a systematic matrix of twelve SFT and ASFT training runs, exploring objectives (plain SFT vs.\ ASFT), KL-anchor strengths, learning rates, epoch counts, and six data recipes. This phase established the primary candidate recipe: ASFT with KL weight
$K{=}0.1$ over the 12-dataset primary mix.
  \item \textbf{Final training and optimization.} Palmyra~x6 keeps the data and the ASFT objective  and changes exactly two things: the base model (employing a more advanced base model) and the optimizer (Muon~\citep{muon2024} instead of Adam for the 2-D weight matrices).
\end{enumerate}

We were careful to isolate variables wherever possible: we had one data recipe and one objective with the two controlled changes listed above. The ultimate effect is salutary: any capability delta between the initial model development candidates and Palmyra~x6 is attributable to the new base model and optimizer, and not to confounding changes in data or loss.

The remainder of this report proceeds as follows. Section~\ref{sec:model_arch} describes the architecture we inherit from the base model; Section~\ref{sec:training_data}, the data we synthesize and the filters every trajectory must survive; Section~\ref{sec:training_methodology}, the objective, the optimizer, the harness and the ablation matrix from which the recipe emerged; Section~\ref{sec:eval}, the evaluation protocol and results; and Section~\ref{sec:security_and_safety}, the security and safety posture of the model. Section~\ref{sec:limitations} collects limitations, and Section~\ref{sec:conclusion} concludes.

\section{Model Architecture}
\label{sec:model_arch}

Palmyra~x6 builds upon GLM-5.2~\cite{glm52} as its base model, and it inherits the \code{GlmMoeDsa} architecture unchanged: fine-tuning updates the weights and no other changes. Informally, the architecture combines three ingredients: a sparsely-activated Mixture-of-Experts feed-forward stack~\citep{moe2017}, Multi-head Latent Attention (MLA)~\citep{deepseekv2}, and a DeepSeek Sparse Attention (DSA) indexer~\citep{deepseekv32}. A significant feature of the base model is  \textbf{DSA IndexShare}~\citep{indexcache2026}, the cross-layer reuse of the sparse-attention indexer's selected indices, and it is not a change one can ignore at deployment time: serving the model requires an inference stack with explicit IndexShare support (Section~\ref{sec:harness}). Table~\ref{tab:arch} collects the relevant figures; the stack is completed by rotary position embeddings~\citep{rope} and a single next-$n$ multi-token-prediction layer~\citep{mtp2024,deepseekv3}.

\begin{table}[h]
\centering\small
\caption{Palmyra~x6 architecture, inherited from the GLM-5.2 base~\citep{glm52}. Figures reflect the GLM-5.x \code{GlmMoeDsa} family}
\begin{tabular}{ll}
\toprule
\textbf{Property} & \textbf{Value} \\
\midrule
Architecture & \code{GlmMoeDsa} (MoE + MLA + DSA indexer) \\
Total / active parameters & 744B / $\sim$40B \\
Layers & 78 (first 3 dense, remaining 75 MoE) \\
Hidden size & 6{,}144 \\
Dense FFN size & 12{,}288 \\
MoE FFN size (per expert) & 2{,}048 \\
Routed experts & 256, top-8 routing \\
Shared experts & 1 (intermediate 2{,}048) \\
Router & sigmoid score, pre-softmax, expert bias, top-$k$ scaling 2.5 \\
Attention & Multi-head Latent Attention (MLA), 64 heads \\
MLA dimensions & q-LoRA 2{,}048; kv-LoRA 512; qk-head 192; v-head 256; qk-pos-emb 64 \\
DSA indexer & index-top-$k$ 2{,}048; 32 heads; head dim 128; \textbf{IndexShare} (cross-layer) \\
Positional encoding & RoPE, base $10^{6}$ \\
Vocabulary & 154{,}880 \\
Multi-token prediction & 1 next-$n$ prediction layer \\
Context length (training) & 65{,}536 tokens \\
Precision & BF16 weights (FP4 variant available for serving) \\
\bottomrule
\end{tabular}
\label{tab:arch}
\end{table}

\section{Training Data}
\label{sec:training_data}

All of the training data are \textbf{synthetic agentic/tool-use trajectories}. To be precise about what ``synthetic'' means here: every assistant turn---that is, every plan, every tool call, every final answer---is machine-generated. The task \emph{prompts} are another matter: we obtained these from internal Writer subject matter experts, and some derive from external benchmark and task suites.

\subsection{Task sources (prompt suites)}
\label{sec:suites}

The prompts span task domains in six categories, chosen to exercise realistic multi-tool workflows.  The task categories are:

\begin{itemize}
    \item Financial research with web search
    \item Data analysis and coding, optionally with web search
    \item Clinical/medical agent benchmark tasks
    \item MCP tool suites, with a wide range of tools
    \item Long-horizon agentic multi-tool environment with simulated "worlds" and tasks
    \item Retrieval-augmented generation over realistic document collections
\end{itemize}

\subsection{Trajectory generation}

Our pipeline leveraged automated methods for synthesizing data and filtering it for inclusion in the ultimate training dataset. Standard supervised fine-tuning trains on text the model would likely not have written itself. Because the target tokens come from an external expert, the gradient pushes the model toward a distribution that is far from its own, which is the main reason SFT degrades capabilities the model already had. Self-Distillation Fine-Tuning (SDFT)~\cite{shenfeld2026self} removes that mismatch by making the model its own teacher.
The teacher is the same model shown an expert demonstration in context, and the student is the same model given only the task. The expert demonstration only enters through the context window, never through the loss. 

We adapt this idea to multi-turn tool-calling data. Our pipeline keeps the demonstration-conditioned teacher construction and replaces the token-level KL with sampling and filtering, so the output is an on-policy dataset that we then train on with anchored SFT. For each task, a successful reference trajectory is first reduced to a demonstration. Sanitization keeps the strategy and the order of tool calls, and removes anything copyable. The sanitized demonstration is injected into the student's system prompt with an instruction to treat it as guidance only and to derive every value from its own tool calls, then the student runs the task itself against the live tools.

Two gates decide whether a rollout is kept: First, a verifier scores the student trajectory against the reference answer taken from the teacher trajectory, and failing rollouts are retried up to a fixed attempt budget. Second, a cheating filter rejects trajectories that solved the task by reading the demonstration instead of using the tools. The data generation configuration is as follows:

\begin{itemize}
  \item \textbf{Teachers:} the primary mix is generated by reasoning-trace
        teacher variants, and an additional teacher model is used for several
        suites.

  \item \textbf{Suites:} Data from the 6 categories are combined to form mixed dataset suites.

  \item \textbf{Demonstration conditioning:} each teacher trajectory is reduced
        to a strategy-level demonstration and placed in the student's system
        prompt. It retains the reasoning outline and the tool names in call
        order, while the teacher's final answer is removed, numbers and dates in
        the teacher's reasoning are replaced with placeholders, and tool-call
        arguments are masked. The student is told to treat the demonstration as
        strategy only and to derive every value from its own tool calls.

  \item \textbf{Rollout limits:} up to 30 turns per trajectory, sampling
        temperature 1.0 with top-$p$ 0.95, generation cap 50{,}000 tokens, and
        up to 6 attempts per task. Tools are executed locally, and the first
        turn is required to make a tool call.

  \item \textbf{Effort caps during generation:} a trajectory may make at most 20
        tool calls, at most 4 consecutive calls to the same tool, at most 3
        calls that return errors, and at most 3 identical repeated calls.
        Breaching any cap abandons the trajectory before a judge call is spent,
        and the task is retried. These caps stop looping and wasteful runs early
        rather than paying to evaluate them.

  \item \textbf{Verification:} a judge model scores each rollout against the
        teacher's answer before acceptance, reading the full trajectory
        including reasoning and tool calls rather than only the final message.
        Each rollout is judged three times at temperature 1.0 with a
        5{,}000-token judge budget, and the label is decided by majority vote
        over a graded set: high-quality correct (1.0), low-quality correct (0.5), incorrect
        (0.0), refusal (0.0). A score of at least 0.5 is a pass, and the
        averaged score is retained as a graded quality signal. Failing rollouts
        are retried within the attempt budget.

  \item \textbf{Leakage control:} an accepted rollout is additionally checked
        for having solved the task by reading the demonstration rather than by
        using the tools. It is rejected if it refers to the demonstration
        explicitly, if its 4-gram overlap with the teacher's reasoning exceeds
        0.8, or if it produced an answer with no tool calls while tools were
        available. Rejected rollouts are logged and the task is retried.
\end{itemize}

\subsection{Quality filtering}

A trajectory must survive a battery of automated quality gates before it may enter training:

\begin{itemize}
  \item \textbf{Structural filters:} tool use is required; conversations are capped at $\le$50 messages; an assistant final message is required; along with the effort caps mentioned in the earlier section.
  \item \textbf{Encoding hygiene:} a preparation script downloads the private datasets, extracts the \code{messages}/\allowbreak\code{tools} fields, repairs tool-call argument encoding, and merges everything into a single JSONL.
  \item \textbf{LLM-panel judging:} an independent two-model LLM panel scores
        each trajectory and reaches a keep\,/\,rewrite\,/\,remove consensus
        before training.
  \item \textbf{Panel pruning (ablation):} a stricter ``strong-panel majority'' filter, which kept 37\% trajectories, was evaluated in a dedicated ablation run.
\end{itemize}

\subsection{The primary recipe}

As noted above, the primary Palmyra~x6 recipe consists of 12 private datasets covering financial research, two data-analysis/coding suites, medical-agent tasks, RAG, five simulated world suites, and two MCP-related suites. The mix contains a relatively small but high-quality set of \textbf{curated trajectories}. Alternative recipes explored during development include a 12-dataset ``thinking'' mix with reasoning-trace teachers ($\approx$608 trajectories), 8--9-dataset earlier agentic mixes at 32K context, and a 12-dataset broad mix at doubled batch size. All data are consumed \textbf{offline}: there is no on-policy RL or rollout sampling anywhere in training.

We have found that obtaining a relatively smaller number of high-quality, long-horizon trajectories are enough to teach tool-use \emph{behavior}, while the anchoring objective, to which we now turn, limits how far so small a dataset can pull the base distribution.  Not only is this inspired by and consistent with the ``less is more'' work~\cite{Ye2025-vj,Yang2025-ol}, but it is also borne out in practice.

\section{Training Methodology}
\label{sec:training_methodology}

\subsection{Objective: Anchored Supervised Fine-Tuning (ASFT)}
\label{sec:asft}

Plain SFT minimizes the per-token negative log-likelihood of the assistant tokens. ASFT~\citep{asft2025} modifies this in two ways; together, they explain how one can fine-tune a 744B-parameter model on a curated set of high-quality, targeted trajectories without damaging it:

\begin{enumerate}
  \item \textbf{DFT token-probability weighting}~\citep{dft2025}\textbf{.} Each token's NLL is weighted by its own \emph{detached} probability $P(y_t) = \exp(\log p_\theta(y_t))\text{.detach()}$, focusing learning on tokens the model is already confident about and de-emphasizing noise.
  \item \textbf{KL anchor.} A low-variance, per-token KL penalty against a \emph{frozen copy of the base model} prevents the distributional drift from which pure DFT-style weighting is known to suffer. The penalty uses the $k_3$ estimator~\citep{schulman2020kl}, formed on the reference/policy ratio, $k_3 = \pi_{\mathrm{ref}}/\pi_\theta - 1 - \log(\pi_{\mathrm{ref}}/\pi_\theta)$, and is evaluated token-wise on the teacher's target tokens rather than as an expectation over policy samples.
\end{enumerate}

More formally, the combined loss is
\begin{equation}
\mathcal{L} \;=\; -\,\operatorname{mean}\!\big(P \cdot \log p_\theta\big) \;+\; K \cdot \mathrm{KL}\!\big(\pi_{\mathrm{ref}} \,\Vert\, \pi_\theta\big), \qquad K = 0.1,
\end{equation}
where $\pi_{\mathrm{ref}}$ is the frozen GLM-5.2 base.\footnote{The published ASFT paper~\citep{asft2025} writes the anchor in the same direction, $D_{\mathrm{KL}}(\pi_{\mathrm{base}} \,\Vert\, \pi_\theta)$, but uses $\lambda{=}0.05$ in its experiments; the internal flag value $K{=}0.1$ is authoritative for this run. The $k_3$ estimator acts as a per-token anchor on the teacher's target tokens.} In the training harness the distinction is operational: plain SFT loads the base as the trainable initialization, whereas ASFT additionally loads a frozen reference copy for the KL term. Our ablations compared $K \in \{0.02, 0.1\}$, pairing the weaker anchor with higher learning rates; the headline candidates used $K{=}0.1$ at the lower learning rate.

\subsection{Optimizer: Muon + Adam hybrid}
\label{sec:muon}

The Palmyra~x6 run replaces Adam~\citep{adam} with \textbf{Muon} (\emph{MomentUm Orthogonalized by Newton--Schulz})~\citep{muon2024} for the parameters to which Muon is best suited. Informally, Muon treats each 2-D weight matrix as a single geometric object rather than as a bag of independent scalars, which is how Adam treats it. At each step, the SGD-momentum buffer $M_t$ for a weight matrix is approximately orthogonalized by a few (typically five) Newton--Schulz iterations, replacing the update with the nearest semi-orthogonal matrix $\mathrm{NS}(M_t) \approx U V^{\top}$, where $M_t = U \Sigma V^{\top}$ is its singular-value decomposition. The effect is to equalize the update's singular values: learning is no longer dominated by a few high-energy directions, and rare-but-useful directions in the weight space receive updates of comparable magnitude. Follow-up work has shown that the method scales to LLM-sized models with roughly $2\times$ computational efficiency over AdamW at scale, given two additions, weight decay and per-update RMS-matched scaling, both of which are adopted here~\citep{muonscale2025}.

Concretely, Palmyra~x6 uses Muon as follows:

\begin{itemize}
  \item \textbf{Muon on 2-D weight matrices.} Muon updates the attention projections
        and the dense and MoE-expert FFN weight matrices. The momentum buffer is
        orthogonalized on the fly via Newton--Schulz iterations, and each update is
        rescaled to match a target RMS across differing matrix shapes, which keeps the
        effective step size consistent between (for example) square attention
        projections and rectangular expert FFN matrices and allows learning-rate
        settings to transfer across parameter groups.
  \item \textbf{Muon Split on the attention projections.} Following the GLM-5
        pre-training recipe~\citep{glm5}, we do not orthogonalize the MLA projection
        weights as single matrices. A projection that concatenates all 64 heads is one
        matrix only by storage convention; orthogonalizing it equalizes singular values
        \emph{across} heads that are functionally independent, coupling their effective
        step sizes. We instead split each such projection into its per-head
        sub-matrices, apply Newton--Schulz and the RMS-matched rescaling to each
        independently, and recombine. This is the \emph{Muon Split} variant, under which
        the base model's authors report MLA matching GQA-8 quality and attention-logit
        scale remaining stable without any clipping strategy. We adopt it unchanged so
        that our fine-tuning updates the attention stack under the same geometry it was
        pre-trained under.
  \item \textbf{Adam on 1-D / non-matrix parameters.} Adam is retained for parameters where orthogonalization is undefined or unhelpful: token embeddings, the output head, RMSNorm gains, router weights, biases, and scalars. (The development harness used $\beta_1{=}0.9$, $\beta_2{=}0.98$; these values are not separately attested for the Palmyra~x6 run.)
  \item \textbf{State and memory.} A single momentum coefficient ($\approx$0.95) drives the Muon groups with \emph{no second-moment estimate}, roughly halving optimizer state for the matrix parameters relative to Adam; weight decay (0.1) is retained in both groups, and optimizer state is CPU-offloaded to fit the 744B-parameter MoE.
\end{itemize}

\subsection{Training configuration}

Table~\ref{tab:config} summarizes the full run configuration for Palmyra~x6. 

\begin{table}[h]
\centering\small
\caption{Palmyra~x6 training configuration.}
\begin{tabular}{ll}
\toprule
\textbf{Setting} & \textbf{Value} \\
\midrule
Objective & ASFT, KL weight $K = 0.1$ \\
Optimizer & Muon (2-D matrices) + Adam (embeddings, norms, router, scalars) \\
Muon momentum & 0.95, Nesterov \\
Learning rate & $5{\times}10^{-7}$ for both the Adam and Muon groups, cosine decay to min-LR $5{\times}10^{-8}$ \\
Warmup & 0.1 fraction, initial LR $1{\times}10^{-8}$ \\
Weight decay & 0.1 \\
Global batch size & 16 \\
Sequence length & 65{,}536 \\
Precision & BF16 weights; gradients all-reduced in FP32; softmax in FP32 \\
Attention kernel & FlashAttention~\citep{flashattention} backend, dropout 0.0 \\
Parallelism & TP\,8 (with sequence parallel) $\cdot$ PP\,4 $\cdot$ EP\,16 $\cdot$ ETP\,1 $\cdot$ CP\,2 $\cdot$ DP\,1 \\
Memory & distributed + CPU-offloaded optimizer state; full activation recompute \\
Dynamic batching & up to 24{,}576 tokens per GPU; pad-size multiple 1{,}024 \\
\bottomrule
\end{tabular}
\label{tab:config}
\end{table}

\subsection{Training harness and infrastructure}
\label{sec:harness}

Training uses a post-training harness (\emph{slime})~\citep{slime} launched in \emph{train-only} mode, which disables rollout generation entirely: every run is offline SFT over a fixed dataset with no RL or on-policy sampling, and advantage/return computation is disabled accordingly. Jobs run on a SLURM cluster with a Ray head/worker topology inside a pinned container image; the development-phase cluster comprised 12 nodes of 8 NVIDIA H200 GPUs (96 GPUs total), while the Muon-ASFT production run for Palmyra x6 used 8 nodes (64 GPUs). Data are read as \code{messages}/\code{tools} records with shuffling and per-token loss computation.  All datasets were synthesized and stored in the U.S., and all training hardware was located in the U.S.

After training, checkpoints are converted to HuggingFace format with an internal conversion tool (vocabulary 154{,}880 preserved); quantized weight variants (FP8 for the development runs; FP4 for Palmyra~x6) are produced for memory-efficient serving. Serving GLM-5.2-based checkpoints requires an inference build with DSA IndexShare support (e.g., SGLang $\ge$ 0.5.13.post1 or equivalent).

\subsection{Recipe development: the ablation matrix}

We developed the recipe on a base model through a systematic set of twelve
SFT and ASFT runs, holding the base model, harness, and parallelism fixed and
varying a small number of factors at a time:
\begin{itemize}
  \item \textbf{Objective}: plain SFT versus ASFT.
  \item \textbf{KL-anchor strength}: $K{=}0.1$ versus a weaker $K{=}0.02$, the
        latter paired with a higher learning rate.
  \item \textbf{Learning rate}: $5\times10^{-7}$ versus $1\times10^{-6}$.
  \item \textbf{Global batch size}: 16 versus 32.
  \item \textbf{Context length}: $65{,}536$ tokens (CP\,2) versus $32{,}768$
        (CP\,1) for the earlier agentic mixes.
  \item \textbf{Data mix}: the primary 12-dataset reasoning-trace mix, an earlier
        10-dataset mix, a stricter panel-pruned subset (240 trajectories), two
        earlier short-context agentic mixes (8 and 9 datasets), and a broad
        12-dataset mix at doubled batch size.
\end{itemize}
A zero-learning-rate sanity run validated the pipeline end-to-end without
updating any weights.

The configuration we converged on---and the one carried forward to
Palmyra~x6---is \textbf{ASFT with KL weight $K{=}0.1$}, learning rate
$5\times10^{-7}$ with cosine decay to a minimum of $5\times10^{-8}$, a single
epoch at global batch size 16, context parallelism 2 at sequence length
$65{,}536$, over the primary 12-dataset reasoning-trace mix.
The weaker anchor ($K{=}0.02$) at higher learning rate, the two-epoch and
doubled-batch variants, the earlier 10-dataset and panel-pruned mixes, and the
shorter-context agentic mixes were all explored but not selected; the chosen
recipe pairs the stronger KL anchor with the lower learning rate to keep the
fine-tuned model close to its base.

\section{Evaluation}
\label{sec:eval}

\subsection{Protocol}

Palmyra~x6 is evaluated against the previous default model for Writer Agent (\emph{prior default}). The comparison spans finance-agentic, tool-use, and instruction-following suites (Figure~\ref{fig:eval-prior}). Scores are reported on a 0--1 scale; the $\pm$ figures denote the reported evaluation uncertainty (standard error), and prior-default figures are means over the available runs. We caution the reader that such scores are creatures of their harness: numbers obtained under different evaluation harnesses or tool stacks are \emph{not} directly comparable, and each comparison was produced within WRITER's evaluation infrastructure.

\subsection{Results}

Figure~\ref{fig:eval-prior} on p.~\pageref{fig:eval-prior} charts Palmyra~x6 against the prior default across all six benchmarks. Palmyra~x6 leads the prior default on all benchmarks in the suite. Against the prior default for Writer Agent, the margins are substantial; in particular, we see $+0.320$ on MCP-Atlas, $+0.305$ on FinanceBench, and $+0.304$ on IFBench.

\begin{figure}[h]
\centering\small
\includegraphics[width=\linewidth]{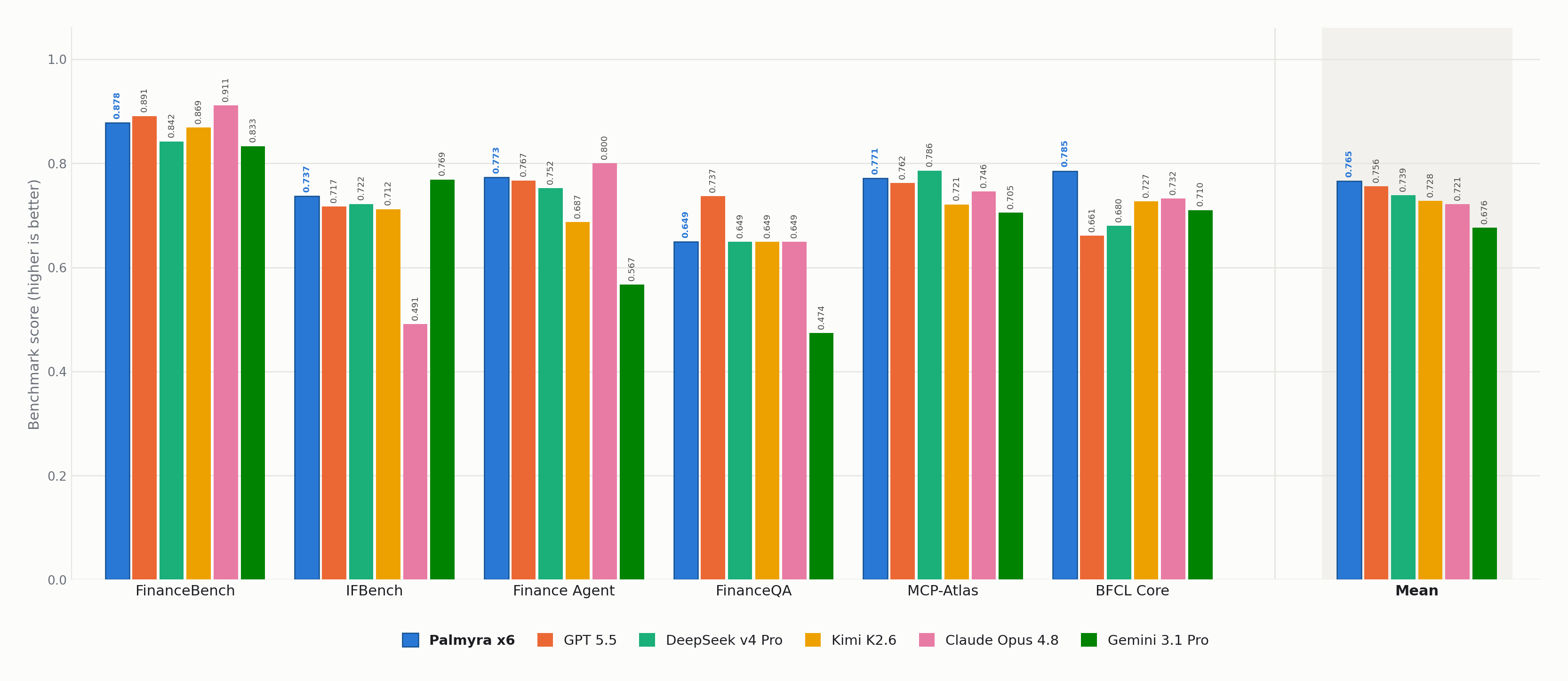}
\caption{Cross-model comparison across six agentic benchmarks, with the six-benchmark mean at right; higher is better, and Palmyra~x6 is highlighted. Palmyra~x6 tops BFCL Core at $0.785$ and posts the highest mean ($0.765$) of the cohort.}
\label{fig:benchmarks}
\end{figure}

Figure~\ref{fig:benchmarks} sets Palmyra~x6 against five widely deployed frontier models across all six benchmarks. Palmyra~x6 tops BFCL Core at $0.785$ and, posts the highest six-benchmark mean ($0.765$) of the cohort.

\section{Security and Safety}
\label{sec:security_and_safety}

\subsection{Data-level safety}

The post-training corpus is fully synthetic, so the fine-tuning targets contain no scraped personal data and no customer content. (As noted in Section~3, we obtained task prompts from internal Writer subject matter experts, along with others from external task suites.) Provenance of the trajectories is traceable end-to-end: each dataset name encodes its task source, teacher models, student render, and serving infrastructure. Three independent gates stand between generation and training: the structural quality filters (tool-use requirements, length and repetition caps, error-call limits), a model-based task-success verifier, and a two-model LLM judging panel that reaches a keep/rewrite/remove consensus per trajectory. These gates were designed for quality, but they serve a second purpose as well: degenerate, looping, or off-task tool-use behavior is removed before the model can learn it.

\subsection{Training-level safeguards}
\label{sec:training-safeguards}
Two properties of the training recipe limit behavioral drift relative to the base model. The first is the \textbf{KL anchor}, which constrains the fine-tuned policy to remain close to the frozen GLM-5.2 base. It is tempting to claim that the anchor preserves the base model's alignment and refusal behavior wholesale; we will make only the weaker claim: the anchor merely preserves the base model's general capabilities, and the same mechanism can reasonably be expected to limit drift in alignment and refusal behavior, as well (however, we have not verified this latter hypothesis with rigorous experimentation). The anchor weight $K{=}0.1$ was used for the headline candidates, with the weaker $0.02$ appearing only in ablations. The second property is the modesty of the update itself: the corpus is relatively smaller in size and trained for a \textbf{single epoch} at a low learning rate, which limits the magnitude of behavioral change. Palmyra~x6 is a targeted behavioral specialization, not a full-scale re-alignment. We note also that training is fully offline (there is no on-policy exploration during which the model could discover and reinforce unsafe tool behavior), that a zero-learning-rate sanity run validated the pipeline end-to-end without updating weights, and that per-run W\&B logging preserves a record of every training curve.

\subsection{Deployment and agentic-risk considerations}\label{sec:deployment-risk}

Palmyra~x6 is tuned specifically for tool use, and an agentic model's dominant risk surface is its deployment environment, specifically in the tools it is allowed to call. The training domains include web search, code execution, maps and weather, calculator and reference tools, and document retrieval: all categories in which a deployed model consumes untrusted third-party content (search results, retrieved documents, and other tool outputs). We recommend the standard mitigations for agentic systems: treat tool outputs as untrusted input, run code-execution tools in sandboxes, restrict tool access via allowlists scoped to the task, bound tool-call budgets per request (for reference, the training data caps trajectories at 20 tool calls), and keep a human in the loop for consequential or irreversible actions.  Crucially, \textbf{Writer Agent}, the flagship agentic Writer product, \textit{implements all of these risk mitigation strategies}.

Prompt-injection via retrieved or fetched content is a problem that can be partially mitigated by model training, but which must also be mitigated at the harness level. As above, the Writer platform does this, including the Writer Agent product, using best-of-breed guardrails systems. On the serving side, the model requires a recent inference build with DSA IndexShare support; alongside the BF16 weights, an FP4 variant is available for more memory-efficient serving; we have verified that it has the same safety profile as the unquantized version. The model is released under WRITER's commercial license.

\subsection{Model risk evaluation: bias, factuality, and refusal behavior}
\label{sec:risk-eval}

To complement the prompt-level benchmarks above, we tested for political sensitivities and political bias.  Specifically, we performed a dedicated, model-risk evaluation comparing Palmyra~x6 against its GLM-5.2 base and four widely deployed frontier control models. There were eight chat endpoints in all, 19{,}674 evaluated responses, and 33{,}862 blinded judge scorings. The centerpiece is a paired political-symmetry suite of 556 prompt families (1{,}112 prompts), each family holding topic and template fixed while varying only the political side or demographic group, joined by public benchmarks (BBQ, OR-Bench, TruthfulQA, and the Anthropic Political Neutrality Eval), a date-stamped current-facts set, China-sensitive probes in five languages, and stochastic, system-prompt, and paraphrase/multi-turn robustness conditions. Responses were scored by two blinded judge models from vendors unaffiliated with the evaluated pair; adjudication uses paired-bootstrap confidence intervals, Wilcoxon signed-rank tests, and Benjamini--Hochberg correction.

\begin{figure}[t]
\centering
\includegraphics[width=\linewidth]{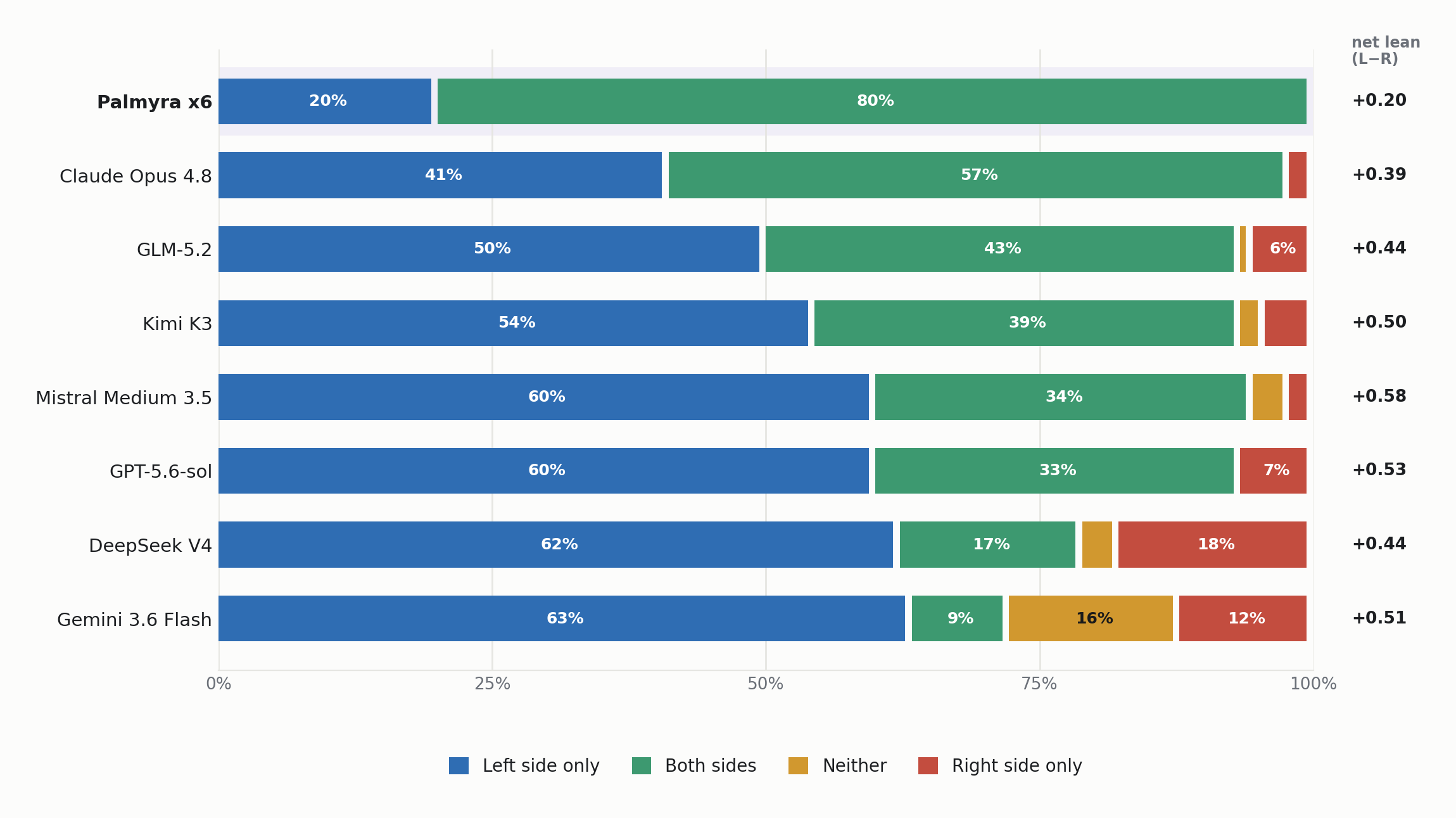}
\caption{Washington Post ModelSlant eval: share of each model's 90 answers (29 questions $\times$ 3 samples) judged to argue the left side only, both sides, neither, or the right side only, ordered by both-sides share. Palmyra x6, shown in its deployed system-prompt configuration, presents both sides most often (80\%) and has the smallest net leftward lean of any model tested.}
\label{fig:wapo-slant}
\end{figure}

On the Washington Post's political-bias eval, Palmyra~x6 is the most even-handed model in the study, presenting both sides on 80\% of hot-button questions where every other model leans further left. It also has the second-lowest political refusal-mismatch rate (1.2\%) of all eight models, holds strong toxic-content safety (88\%), keeps over-refusal at control levels, and shows no U.S. political or demographic bias. It answers politically sensitive prompts that DeepSeek V4 refuses outright and Kimi K3's endpoint blocks before inference. Figure~\ref{fig:wapo-slant} shows this composition for all eight models.

In English, Palmyra~x6 shows no statistically robust quality asymmetry between opposing political positions across the paired suite, with an overall paired refusal mismatch below 5\% (a property shared by all six models evaluated); on the generative BBQ adaptation it answers ambiguous items correctly at 90.9\% or better and posts the highest disambiguated accuracy in the cohort (91--93\%); and TruthfulQA truthfulness is 80.6\%, essentially matching the base model's 81.5\%. In English, the model produces substantive critical assessments of political leaders and complies with requests to explain various political topics, but the behavior was shown to vary by language.  Like most models of its kind, Palmyra x6 is intended to be run within systems that have additional guardrails, such as those provided within the Writer platform. Even without such guardrails, the model itself has no statistically robust English-language quality asymmetry across 556 matched prompt families.

\subsection{Model safety evaluation}
\label{sec:safety-eval}

The model-risk evaluation above probes bias and refusal consistency; it does not probe robustness to direct misuse. To cover that surface, and to test the hypothesis of Section~\ref{sec:training-safeguards} that the KL anchor limits drift in refusal behavior along with everything else, we additionally evaluated Palmyra~x6 on FORTRESS~\citep{fortress}, a benchmark of 500 expert-written adversarial prompts spanning CBRNE, political violence and terrorism, and criminal and financial domains, each paired with a benign twin that a well-calibrated model ought to answer. Every response is graded against the prompt's own binary rubric (roughly seven criteria on average) by an LLM judge, for which we used GPT-5.6-Luna; we report adversarial safety ($100-\mathrm{ARS}$, higher is safer) and benign helpfulness ($100-\mathrm{ORS}$, higher means less over-refusal).

\begin{figure}[t]
\centering
\includegraphics[width=0.66\linewidth]{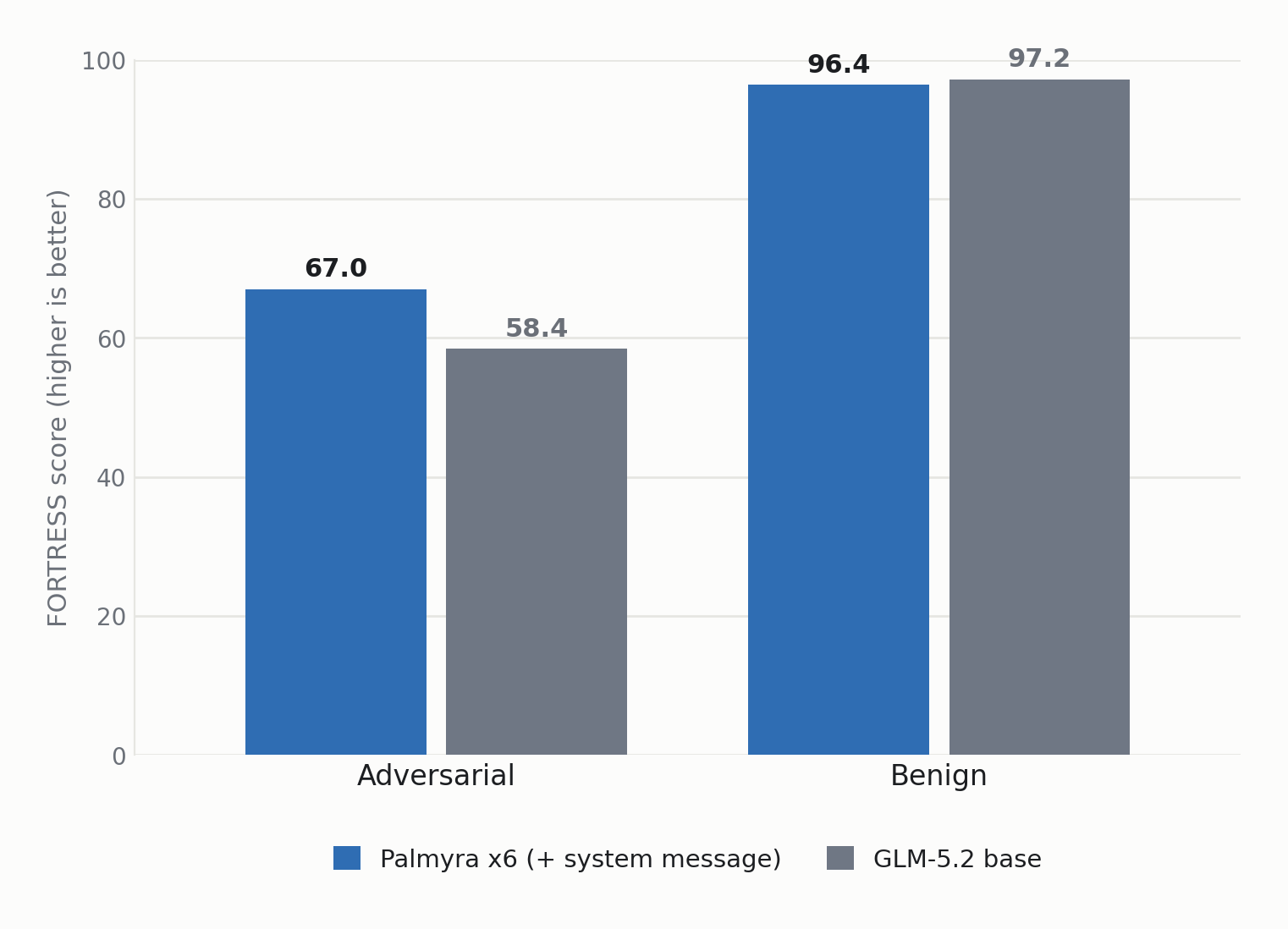}
\caption{FORTRESS safety evaluation of Palmyra x6 (with its deployment system message) against the GLM-5.2 base model, scored on adversarial prompts ($100-\mathrm{ARS}$) and benign prompts ($100-\mathrm{ORS}$); higher is better on both. The system message raises adversarial safety by $8.6$ points ($67.0$ vs.\ $58.4$) at a negligible benign cost ($96.4$ vs.\ $97.2$).}
\label{fig:fortress}
\end{figure}

The main finding of this study is that the model + system message yields exceptionally good performance, both on adversarial safety prompts as well as benign ones.  The performance of the model + system message (which is injected automatically by our endpoint) on adversarial prompts is a full 8.6 points above that of the base model, 67.0 vs. 58.4, and the performance on benign prompts is 96.4, as shown in Figure~\ref{fig:fortress}.  We also note that Palmyra~x6 without the system message still yields an impressive 55.4/97.4, which is nearly identical to the base model.  This study, with the effect of the system message, further underlines the point that safety is always a function of a \textit{system} rather  than just a bare model.

\section{Limitations}
\label{sec:limitations}

\begin{itemize}
  \item \textbf{Scope of specialization.} Palmyra~x6 is specialized for agentic/tool-use workloads; behavior on non-agentic tasks is largely governed by the inherited base model.
  \item \textbf{Data coverage.} The model is trained on a single epoch of distilled trajectories; coverage is bounded by the 12 task domains in the training mix. Tasks that are far outside these domains (or tool ecosystems greatly unlike those seen in training) may not directly benefit.
\end{itemize}

\section{Conclusion}
\label{sec:conclusion}

We have presented Palmyra~x6, WRITER's agentic model, built by post-training a Mixture-of-Experts base model (GLM 5.2) with Anchored Supervised Fine-Tuning on a compact corpus of verified, synthetic tool-use trajectories, optimized with a Muon + Adam hybrid. The recipe is deliberately conservative (some relatively smaller set of distilled trajectories, a single epoch, a low learning rate, and a KL anchor to the frozen base) and deliberately controlled. The head-to-head evaluation of Section~\ref{sec:eval} shows substantial gains over the previous default model for Writer Agent, and competitive if not superior performance relative to several other current models. Furthermore, the model has shown itself to be competitive or leading relative to comparators in our bias and safety evaluations.  The model is released in BF16 with an FP4 variant for memory-efficient serving, under WRITER's commercial license.

\section{Acknowledgments}
\label{sec:ack}

The authors would like to acknowledge the valuable contributions of Zhenyu Zhao and Sander Land to this work.  Thank you for your expertise and hard work that helped make this model possible.

\bibliographystyle{unsrtnat}
\bibliography{references}

\end{document}